\documentclass[10pt,twocolumn,letterpaper]{article}

\usepackage{cvpr}
\usepackage{times}
\usepackage{epsfig}
\usepackage{graphicx}
\usepackage{amsmath}
\usepackage{amssymb}

\usepackage{bm}
\usepackage{algorithm}  
\usepackage{algorithmicx} 
\usepackage{algpseudocode} 

\usepackage[pagebackref=true,breaklinks=true,letterpaper=true,colorlinks,bookmarks=false]{hyperref}

\cvprfinalcopy 

\def\cvprPaperID{6430} 

\ifcvprfinal\fi
\begin{document}

\title{Learning from Web Data with Self-Organizing Memory Module}

\author{Yi Tu, Li Niu\thanks{Corresponding author.}, Junjie Chen, Dawei Cheng, and Liqing Zhang\footnotemark[1]\\
	MoE Key Lab of Artificial Intelligence, Department of Computer Science and Engineering\\
	Shanghai Jiao Tong University, Shanghai, China\\
{\tt\small \{tuyi1991,ustcnewly,chen.bys,dawei.cheng \}@sjtu.edu.cn, zhang-lq@cs.sjtu.edu.cn}}

\maketitle

\begin{abstract}
Learning from web data has attracted lots of research interest in recent years. 
However, crawled web images usually have two types of noises, label noise and background noise, which induce extra difficulties in utilizing them effectively.
Most existing methods either rely on human supervision or ignore the background noise.
In this paper, we propose a novel method, which is capable of handling these two types of noises together, without the supervision of clean images in the training stage. 
Particularly, we formulate our method under the framework of multi-instance learning by grouping ROIs (i.e., images and their region proposals) from the same category into bags. ROIs in each bag are assigned with different weights based on the representative/discriminative scores of their nearest clusters, in which the clusters and their scores are obtained via our designed memory module. 
Our memory module could be naturally integrated with the classification module, leading to an end-to-end trainable system. 
Extensive experiments on four benchmark datasets demonstrate the effectiveness of our method.
\end{abstract}

\section{Introduction}
\begin{figure}[tb]
    \centering
    \includegraphics[width=3.2in]{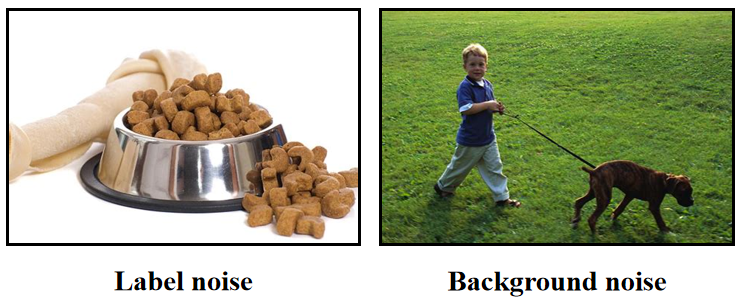}
    \caption{Two web images crawled with keyword ``dog". \textbf{Left:} Dog food; \textbf{Right:} A kid and a dog on the grassland.}
    \label{noises}
\end{figure}

Deep learning is a data-hungry method that demands large numbers of well-labeled training samples, but acquiring massive images with clean labels is expensive, time-consuming, and labor-intensive. Considering that there are abundant freely available web data online, learning from web images could be promising. However, web data have two severe flaws: label noise and background noise. Label noise means the incorrectly labeled images. Since web images are usually retrieved by using the category name as the keyword when searching from public websites, unrelated images might appear in the searching results.
Different from label noise, background noise is caused by the cluttered and diverse contents of web images compared with standard datasets. Specifically, in manually labeled datasets like Cifar-10, the target objects of each category usually appear at the center and occupy relatively large areas, yielding little background noise. However, in web images, background or irrelevant objects may occupy the majority of the whole image. 
One example is provided in Figure \ref{noises}, in which two images are crawled with the keyword ``dog''. The left image belongs to label noise since it has dog food, which is indirectly related to ``dog". Meanwhile, the right image belongs to background noise because the grassland occupies the majority of the whole image, and a kid also takes a salient position.

There are already many studies \cite{Niu1,The-Unreasonable,MakeNet,TIP-Benefit,ICCV-wang2019symmetric, ICCV-huang2019o2u, niu22015visual,niu32016exploiting,niu42018webly} on using web images to learn classifiers. However, most of them~\cite{DRAE,CleanNet,CurriculumNet,Niu1,The-Unreasonable,wsl-reweight,Mentornet} only focused on label noise. 
In contrast, some recent works began to consider the background noise.
In particular, Zhuang \etal \cite{Attend} used attention maps to suppress background noise, but this method did not fully exploit the relationship among different regions, which might limit its ability to remove noisy regions. 
Sun \etal \cite{TIP-Benefit} utilized weakly supervised region proposal network to distill clean region proposals from web images, but this approach requires extra clean images in the training stage. 

In this work, we propose a novel method to address the label noise and background noise simultaneously without using human annotation. We first use an unsupervised proposal extraction method \cite{EdgeBoxes} to capture image regions which are likely to contain meaningful objects. In the remainder of this paper, we use ``ROI" (Region Of Interest) to denote both images and their region proposals. Following the idea of multi-instance learning, ROIs from the same category are grouped into bags and the ROIs in each bag are called instances. Based on the assumption that there is at least a proportion of clean ROIs in each bag, we tend to learn different weights for different ROIs with lower weights indicating noisy ROIs, through which label/background noise can be mitigated. With ROI weights, we can use the weighted average of ROI-level features within each bag as bag-level features, which are cleaner than ROI-level features and thus more suitable for training a robust classifier.

Instead of learning weights via self-attention like~\cite{ilse2018attention,Attend}, in order to fully exploit the relationship among different ROIs, we tend to learn ROI weights by comparing them with prototypes, which are obtained via clustering bag-level features. Each cluster center (\emph{i.e.}, prototype) has a representative (\emph{resp.}, discriminative) score for each category, which means how this cluster center is representative (\emph{resp.}, discriminative) for each category. Then, the weight of each ROI can be calculated based on its nearest cluster center for the corresponding category. Although the idea of the prototype has been studied in many areas such as semi-supervised learning~\cite{MA_DNN} and few-shot learning~\cite{snell2017prototypical}, they usually cluster the samples within each category, while we cluster bag-level features from all categories to capture the cross-category relationship.

Traditional clustering methods like K-means could be used to cluster bag-level features. However, we use recently proposed key-value memory module~\cite{Key-Value-Memory} to achieve this goal, which is more powerful and flexible. The memory module could be integrated with the classification module, yielding an end-to-end trainable system. Moreover, it can online store and update the category-specific representative/discriminative scores of cluster centers at the same time. As a minor contribution, we adopt the idea of Self-Organizing Map \cite{SOM} to improve the existing memory module to stabilize the training process. 

Our contributions can be summarized as follows: 1) The major contribution is handling the label/background noise of web data under the multi-instance learning framework with the memory module; 2) The minor contribution is proposing the self-organizing memory module to stabilize the training process and results; 3) The experiments on several benchmark datasets demonstrate the effectiveness of our method in learning classifiers with web images. 

\section{Related Work}
\subsection{Webly Supervised Learning}
For learning from web data, previous works focus on handling the label noise in three directions, removing label noise \cite{A-singh2012unsupervised,B-dai2013ensemble,C-dai2016unsupervised,I-misra2016seeing,DRAE,CleanNet,Niu1,The-Unreasonable,wsl-reweight,7-survey,re-ren2018learning,ICCV-han2019deep}, building noise-robust model \cite{D-chen2013neil,E-divvala2014learning,F-chen2014enriching,MakeNet,label-flip,label-flip-wsl,bootstrapping,re-vahdat2017toward,bs-tanaka2018joint, ICCV-kim2019nlnl}, and curriculum learning \cite{CurriculumNet,Mentornet}.
The above approaches focused on label noise. However, web data also have background noise, as mentioned in \cite{TIP-Benefit}. To address this issue,
Zhuang \etal \cite{Attend} utilized the attention mechanism \cite{Attention} to reduce the attention on background regions while
Sun \etal \cite{TIP-Benefit} used a weakly unsupervised object localization method to reduce the background noise. 

Most previous works utilize extra information like a small clean dataset or only consider the label noise issue. In contrast, our method can solve both label noise and background noise by only using noisy web images in the training stage. 
\begin{figure*}[htb]
    \centering
    \includegraphics[width=6.5in]{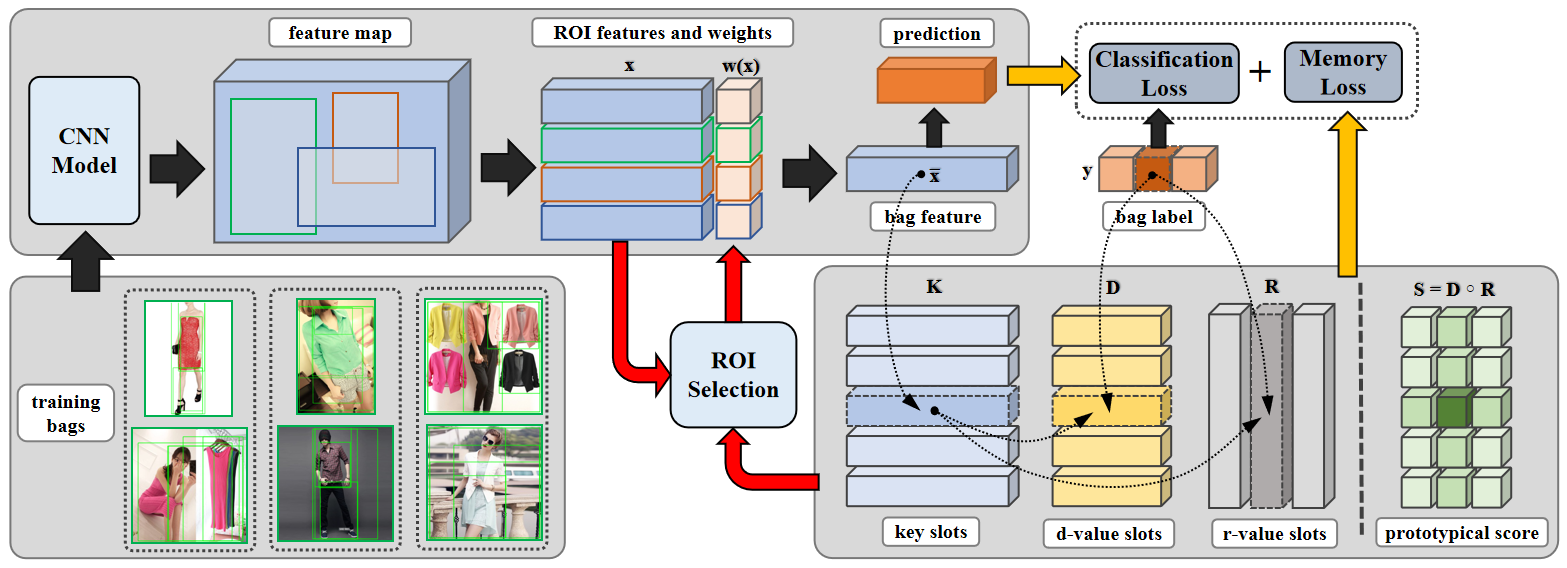}
    \caption{Illustration of our method. \textbf{Black Arrow:} Update the CNN model with bag-level features and bag labels. \textbf{Dashed Arrow:} Update the memory module with bag-level features and bag labels. \textbf{Red Arrow:} Update the weights of ROIs based on the memory module. The whole training algorithm is listed in Algorithm~\ref{alg1}.} 
    \label{model}
\end{figure*}
\subsection{Memory Networks}
Memory networks were recently introduced to solve the question answering task \cite{Memory-networks,End-To-End-Memory,Key-Value-Memory}. 
Memory network was first proposed in~\cite{Memory-networks} and extended to be end-to-end trainable in \cite{End-To-End-Memory}.
Miller \etal \cite{Key-Value-Memory} added the key and value module for directly reading documents, rendering the memory network more flexible and powerful.
More recently, memory networks have been employed for one-shot learning \cite{MetaMN,rare-events}, few-shot learning \cite{xu2017few,CMN}, and semi-supervised learning \cite{MA_DNN}.

Although memory networks have been studied in many tasks, our work is the first to utilize the memory network to handle the label/background noise of web data.

\subsection{Multi-Instance Learning}
In multi-instance learning (MIL), multiple instances are grouped into a bag, with at least one positive instance triggering the bag-level label. The main goal of MIL is to learn a robust classifier with unknown instance labels. Some early methods based on SVM \cite{sMIL,KI-SVM} treat one bag as an entirety or infer instance labels within each bag. In the deep learning era, various pooling operations have been studied like mean pooling and max pooling \cite{From-image-level,DMIL,Deep-MIML}. Different from these non-trainable pooling operators, some works \cite{WeightedMIL,ilse2018attention,Classifying-segmenting,G-wan2019c} proposed trainable operators to learn different weights for different instances. 
By utilizing the attention mechanism, Pappas and PopescuBelis \cite{WeightedMIL} proposed an attention-based MIL with attention weights trained in an auxiliary linear regression model. 
AD-MIL \cite{ilse2018attention} took a further step and designed permutation-invariant aggregation operator with the gated attention mechanism.

Under the MIL framework, we utilize a memory module to learn weights for the instances in each bag, which has not been explored before.

\section{Methodology}
In this paper, we denote a matrix/vector by using a uppercase/lowercase letter in boldface (\eg, $\mathbf{A}$ denotes a matrix and $\mathbf a$ denotes a vector). 
 $\mathbf{a}_i$ denotes a particular row or column of $\mathbf{A}$ indexed by subscript $i$. $a_{i,j}$
denotes an element of $\mathbf{A}$ at the $i$-th row and $j$-th column. Moreover, we use $\mathbf{a}^T$ to denote the transpose of $\mathbf{a}$, and $\mathbf{A} \circ \mathbf{B}$ to denote the element-wise product between $\mathbf{A} $ and $\mathbf{B}$. 

\subsection{Overview of Our Method}
The flowchart of our method is illustrated in Figure \ref{model}. We first extract region proposals for each image using the unsupervised EdgeBox method \cite{EdgeBoxes}. By tuning the hyper-parameters in EdgeBox, we expect the extracted proposals to cover most objects to avoid missing important information (see details in Section \ref{Implementation-Details}). We group ROIs (\emph{i.e.}, images and their proposals) from the same category into training bags, in which ROIs within each bag are treated as instances. To assign different weights to different ROIs in each bag, we compare each ROI with its nearest key in the memory module. Then we take the weighted average of ROI-level features as bag-level features, which are used to train the classifier and update the memory module.

\subsection{Multi-Instance Learning Framework}\label{sec:weight_init}

We build our method under the multi-instance learning framework. 
Particularly, we group several images of the same category and their region proposals into one bag, so that each bag has multiple instances (\emph{i.e.}, ROIs). We aim to assign higher weights to clean ROIs and use the weighted average of ROI-level features within each bag as bag-level features, which are supposed to be cleaner than ROI-level features.

Formally, we use $\mathcal{S}$ to denote the training set of multiple bags, and $\mathcal{B} \in\mathcal{S}$ denotes a single training bag. Note that we use the same number $n_g$ of images in each bag and generate the same number $n_p$ of region proposals for each image, leading to the same number $n_b$ of ROIs in each bag with $n_b=n_g(n_p+1)$.
Specifically, $\mathcal{B}= \{\mathbf{x}_i |i=1,2,...,n_b\}$ is a bag of $n_b$ ROIs, in which $\mathbf{x}_i\in\mathcal{R}^d$ is the $d$-dim feature vector of $i$-th ROI. We use $\texttt{w}(\mathbf{x}_i)$ to denote the weight of $\mathbf{x}_i$ with $\sum_{\mathbf{x}_i\in\mathcal{B}}\texttt{w}(\mathbf{x}_i) =1$. As shown in Figure \ref{model}, the features of ROIs are pooled from the corresponding regions on the feature map of the last convolutional layer in CNN model, similar to \cite{Faster-RCNN}.

Given a bag with category label $y\in[1,2,...,C]$ with $C$ being the number of total categories, we can also represent its bag label as a $C$-dim one-hot vector $\mathbf{y}$ with only the $y$-th element being one. 
After assigning weight $\texttt{w}(\mathbf{x}_i)$ to each $\mathbf{x}_i$, we use weighted average of ROI features in each bag as the bag-level feature:
$\bar{\mathbf{x} }= \sum_{\mathbf{x}_i\in\mathcal{B}} \texttt{w}(\mathbf{x}_i)\cdot \mathbf{x}_i \in\mathcal{R}^d.$
Our classification module is based on bag-level features with the cross-entropy loss:
\begin{equation} \label{eqn:L_cls}
\mathcal{L}_\text{cls} 
=-\sum_{\mathcal{B}\in\mathcal{S}}
\mathbf{y}^T \log\left(f(\sum_{\mathbf{x}_i\in\mathcal{B}} \texttt{w}(\mathbf{x}_i)\cdot \mathbf{x}_i)\right),
\end{equation}
in which $f(\cdot)$ is a softmax classification layer. At the initialization step, the weights of region proposals are all set as zero while the images are assigned with uniform weights in each bag. We use  $\bar{\texttt{w}}(\mathbf{x}_i)$ to denote such initialized ROI weights.
After initializing the CNN model, we tend to learn different weights for ROIs by virtue of the memory module. Next, we first introduce our memory module and then describe how to assign weights to ROIs based on our memory module.

\subsection{Self-Organizing Memory Module}
\label{sec:memory-module}
The basic function of our memory module is clustering bag-level features and each cluster center can be treated as a prototype~\cite{A-singh2012unsupervised,B-dai2013ensemble,C-dai2016unsupervised,MA_DNN,snell2017prototypical}. 
Although traditional clustering methods like K-means can realize a similar function, the memory module is more flexible and powerful. Specifically, the memory module can be easily integrated with the classification module, leading to an end-to-end trainable system. Besides, the memory module can simultaneously store and update additional useful information, \emph{i.e.}, category-specific representative/discriminative scores of each cluster center. 

\subsubsection{Memory Module Architecture:}

Our memory module consists of key slots and values slots. The key slots contain cluster centers, while the value slots contain their corresponding category-specific representative/discriminative scores. 

We use $\mathbf{K}\in \mathcal{R}^{d\times L}$ to denote key slots, in which $L$ is the number of key slots and the $l$-th column $\mathbf{k}_l$ is the $l$-th key slot representing the $l$-th cluster center.

To capture the relationship between clusters and categories,  we design two value slots. We investigate two types of cluster-category relationships, \emph{i.e.}, how discriminative and how representative a learned cluster is to a category.
Correspondingly, we have two types of value slots: the ``discriminative" value slots (d-value slots) and the ``representative" value slots (r-value slots).
Each pair of  d-value slot and r-value slot corresponds to one key slot.  
Formally, we use $\mathbf{D} \in\mathcal{R}^{C\times L}$ (\emph{resp.}, $\mathbf{R}\in\mathcal{R}^{C\times L}$) to denote d-value (\emph{resp.}, r-value) slots, where $d_{y,l}$ (\emph{resp.}, $r_{y,l}$) is the discriminative (\emph{resp.}, representative) score of  $\mathbf{k}_l$ for the $y$-th category. 

To better explain discriminative/representative score, we assume that  $L$ clusters are obtained based on all training bags and $n_{y,l}$ is the number of training bags from the $y$-th category in the $l$-th cluster. Then, we can calculate $\tilde{d}_{y,l}=\frac{n_{y,l}}{\sum_{y=1}^C n_{y,l}}$ and $\tilde{r}_{y,l}=\frac{n_{y,l}}{\sum_{l=1}^L n_{y,l}}$. Intuitively, $\tilde{d}_{y,l}$ means the percentage of the bags from the $y$-th category among the bags in the $l$-th cluster. The larger $\tilde{d}_{y,l}$ is, the more discriminative the $l$-th cluster is to the $y$-th category. So we expect $d_{y,l}$ in d-value slots $\mathbf{D}$ to approximate  $\tilde{d}_{y,l}$.
Similarly, $\tilde{r}_{y,l}$ means the percentage of the bags in the $l$-th cluster among the bags from the $y$-th category. The larger $\tilde{r}_{y,l}$ is, the more representative the $l$-th cluster is to the $y$-th category. So we expect $r_{y,l}$ in r-value slots $\mathbf{R}$ to approximate $\tilde{r}_{y,l}$.

\subsubsection{Memory Module Updating:}

With all key and value slots randomly initialized, they are updated based on the bag-level feature $\bar{\mathbf{x} }$ of training bag $\mathcal{B}$ from $c$-th category and its one-hot label vector $\mathbf{y}$. 

First, we seek for the cluster that $\bar{\mathbf{x} }$ belongs to, which is also referred to as the ``winner key slot'' of $\bar{\mathbf{x} }$. Precisely, we calculate the cosine similarity between $\bar{\mathbf{x} }$ and all keys as $\text{cos}(\bar{\mathbf{x} }, \mathbf{k}_l)=
\frac{\mathbf{k}_l^T \bar{\mathbf{x} }}{\|\mathbf{k}_l\|_2 \|\bar{\mathbf{x} }\|_2}$ for $l=1,2,...,L$, and find the winner key slot $\mathbf{k}_z$ of $\bar{\mathbf{x} }$ where:
\begin{equation}
z= \arg\max\limits_{l} \text{cos}(\bar{\mathbf{x} }, \mathbf{k}_l)
\end{equation}

After determining that $\bar{\mathbf{x} }$ belongs to the $z$-th cluster, the cluster center $\mathbf{k}_z$ needs to be updated based on $\bar{\mathbf{x} }$. Unlike the previous approach \cite{CMN} which updates cluster centers with computed gradients, we employ a loss function which is more elegant and functionally similar:  
\begin{equation}
\label{equation:key-loss-1}
\mathcal{L}_\text{key}=
-\sum_{\mathcal{B}\in\mathcal{S}}
\text{cos}(\bar{\mathbf{x} }, \mathbf{k}_z),
\end{equation}
which can push the winner cluster center $\mathbf{k}_z$ closer to $\bar{\mathbf{x}}$.

With similar loss functions, we also update d-value slots and r-value slots accordingly.
For d-value slots $\mathbf{D}$, recall that we expect $d_{y,i}$ to approximate $\tilde{d}_{y,i}$ which means the percentage of the bags from the $y$-th category among the bags in the $i$-th cluster. Then, the $z$-th column $\mathbf{d}_z$ of $\mathbf{D}$ could represent the category distribution in the $z$-th cluster, so we need to update $\mathbf{d}_z$ with the label vector $\mathbf{y}$ of $\bar{\mathbf{x} }$ as follows,
\begin{eqnarray} 
&&\mathcal{L}_\text{d-value}
=-\sum_{\mathcal{B}\in\mathcal{S}}
\text{cos}(\mathbf{y}, \mathbf{d}_z)
\label{eqn:d-value-loss},\\
&&\text{ s.t. } \|\mathbf{d}_z\|_1=1,\ \mathbf{d}_z \geq \mathbf{0}.\label{eqn:d-value-loss-st}
\end{eqnarray}
$\mathcal{L}_\text{d-value}$ can push $\mathbf{d}_z$ towards $\mathbf{y}$ while maintaining $\mathbf{d}_z$ as a valid distribution with (\ref{eqn:d-value-loss-st}), so $d_{y,z}$ will approximate $\tilde{d}_{y,z}$ eventually. 

For r-value slots $\mathbf{R}$, recall that we expect $r_{y,i}$ to approximate $\tilde{r}_{y,i}$ which means the percentage of the bags in the $i$-th cluster among the bags from the $y$-th category. Then, $\mathbf{r}_y$, the $y$-th row of $\mathbf{R}$, could represent the distribution of all bags from the $y$-th category over all clusters, so we need to update $\mathbf{r}_y$ with the one-hot cluster indicator vector $\mathbf{z}$ of $\bar{\mathbf{x} }$ (only the $z$-th element is $1$) as follows,
\begin{eqnarray}
&&\mathcal{L}_\text{r-value}
=-\sum_{\mathcal{B}\in\mathcal{S}}
\text{cos}(\mathbf{z}, \mathbf{r}_y)
,\label{eqn:r-value-loss}\\
&&\text{ s.t. }\|\mathbf{r}_y\|_1=1,\,\mathbf{r}_y \geq \mathbf{0}. \label{eqn:r-value-loss-st}
\end{eqnarray}
Similar to $\mathbf{d}_z$, $\mathcal{L}_\text{r-value}$ can push $\mathbf{r}_y$ towards $\mathbf{z}$ while keeping it a valid distribution with  (\ref{eqn:r-value-loss-st}), so $r_{y,z}$ will approximate $\tilde{r}_{y,z}$ eventually. The theoretical proof and more details for $\mathcal{L}_\text{d-value}$ and $\mathcal{L}_\text{r-value}$ can be found in Supplementary. 



\subsubsection{Self-Organizing Map (SOM) Extension:}

A good clustering algorithm should be insensitive to initialization and produce balanced clustering results. Inspired by Self-Organizing Map \cite{SOM}, we design a neighborhood constraint on the key slots to achieve this goal, leading to our self-organizing memory module.   

In particular, we arrange the key slots on a square grid. When updating the winner key slot $\mathbf{k}_z$, we also update its spatial neighbors. The neighborhood of $\mathbf{k}_z$ is defined as $\mathcal{N}(\mathbf{k}_z, \delta) = \{\mathbf{k}_i|\text{geo}(\mathbf{k}_z,\mathbf{k}_i) \leq\delta\}$, in which $\text{geo}(\cdot,\cdot)$ is the geodesic distance of two key slots on the square grid and $\delta$ is a hyper-parameter that controls the neighborhood size. 


Then, the key loss $\mathcal{L}_\text{key}$ in (\ref{equation:key-loss-1}) can be replaced by
 \begin{equation}
\mathcal{L}_\text{SOM-key}=-
\sum_{\mathcal{B}\in\mathcal{S}}
\sum_{\mathbf{k}_i\in \mathcal{N}(\mathbf{k}_z, \delta)}
\eta(\mathbf{k}_z, \mathbf{k}_i) \cdot
\text{cos}(\bar{\mathbf{x} }, \mathbf{k}_i),    
\end{equation}
in which $\eta(\mathbf{k}_z,\mathbf{k}_i) =(1+\text{geo}(\mathbf{k}_z,\mathbf{k}_i))^{-1}$ is the weight assigned to $\mathbf{k}_i$ and  negatively correlated with the geodesic distance (see more technical details in the Supplementary).
In summary, the total loss of updating our self-organizing memory module can be written as:
\begin{eqnarray}\label{eqn:L_mem}
\mathcal{L}_\text{memory}= \mathcal{L}_\text{SOM-key}+\mathcal{L}_\text{d-value}+\mathcal{L}_\text{r-value}.
\end{eqnarray}


\subsection{ROI Selection Based on Memory Module}
\label{sec:MIL}

Based on the memory module, we can assign different weights to different ROIs in each bag. Specifically, given a ROI $\mathbf{x} $ with its bag label $y$, we first seek for its winner key slot $\mathbf{k}_z$, and obtain the discriminative (\emph{resp.}, representative) score of $\mathbf{k}_z$ for the $y$-th category, that is, $d_{y,z}$ (\emph{resp.}, $r_{y,z}$). For a clean ROI, we expect its winner key to be both discriminative and representative for its category. For ease of description, we define $\mathbf{S}=\mathbf{D} \circ\mathbf{R}$ with $s_{y,z} = d_{y,z}\cdot r_{y,z}$. We refer to $s_{y,z}$ as the prototypical score of $\mathbf{k}_z$ for the $y$-th category.
Therefore, ROIs with higher prototypical scores are more prone to be clean ROIs. 

Besides the prototypical score, we propose another discount factor by considering ROI areas.
Intuitively, we conjecture that smaller ROIs are less likely to have meaningful content and thus should be penalized. Thus, we use area score (a-score) $\sigma(\cdot)$ to describe the relative size of each ROI. Recall that there are two types of ROIs in each bag: image and region proposal. For original images, we set $\sigma(\mathbf{x} )=1$.
For region proposals, we calculate $\sigma(\mathbf{x} )$ as the ratio between the area of region proposal $\mathbf{x} $ and the maximum area of all region proposals (excluding the full image) from the same image. 
To this end, we use a-score $\sigma(\mathbf{x} )$ to discount $s_{y,z}$, resulting in a new weight for $\mathbf{x} $: 
 \begin{equation}
  \texttt{w}(\mathbf{x} ) =  s_{y,z}\cdot \sigma(\mathbf{x} ).
 \label{weight}
 \end{equation}

After calculating the ROI weights based on (\ref{weight}),  we only keep the top $p$ (\eg, 10\%) weights of ROIs in each bag while the other weights are set to be zero.
The ROI weights in each bag are then normalized so that they sum to one.

\begin{algorithm}[tb]  
  \caption{\textbf{:} The Training Process of Our Network}  
  \label{alg1}  
  \begin{algorithmic}[1]
\Require  
Bags of ROIs $\mathcal{B}$ and bag label $\mathbf{y}$. Initialize $p=$10\%. Initialize ROI weights as $\bar{\texttt{w}}(\mathbf{x}_i)$ in Section~\ref{sec:weight_init}.
\Ensure  
Model parameters $\{\bm{\theta}_\text{cnn}, \bm{\theta}_\text{mem}\}$.
\State Initialize $\bm{\theta}_\text{cnn}$ based on (\ref{eqn:L_cls}) with $\bar{\texttt{w}}(\mathbf{x}_i)$.
\State Initialize $\bm{\theta}_\text{mem}$ based on (\ref{eqn:L_mem}) with $\bar{\texttt{w}}(\mathbf{x}_i)$.
\State \textbf{Repeat}:
\State \hspace{1em}Update $\bm{\theta}_\text{cnn}$ and $\bm{\theta}_\text{mem}$ based on (\ref{L_total}) while $\texttt{w}(\mathbf{x}_i)$ are updated based on (\ref{weight}) accordingly.
\State \hspace{1em}$p\gets p+5\%$.
\State \hspace{1em}Break if $p>40\%$.
\State \textbf{End Repeat.}
\end{algorithmic}  
\end{algorithm} 

\begin{figure*}[htb]
    \centering
    \includegraphics[width=16cm]{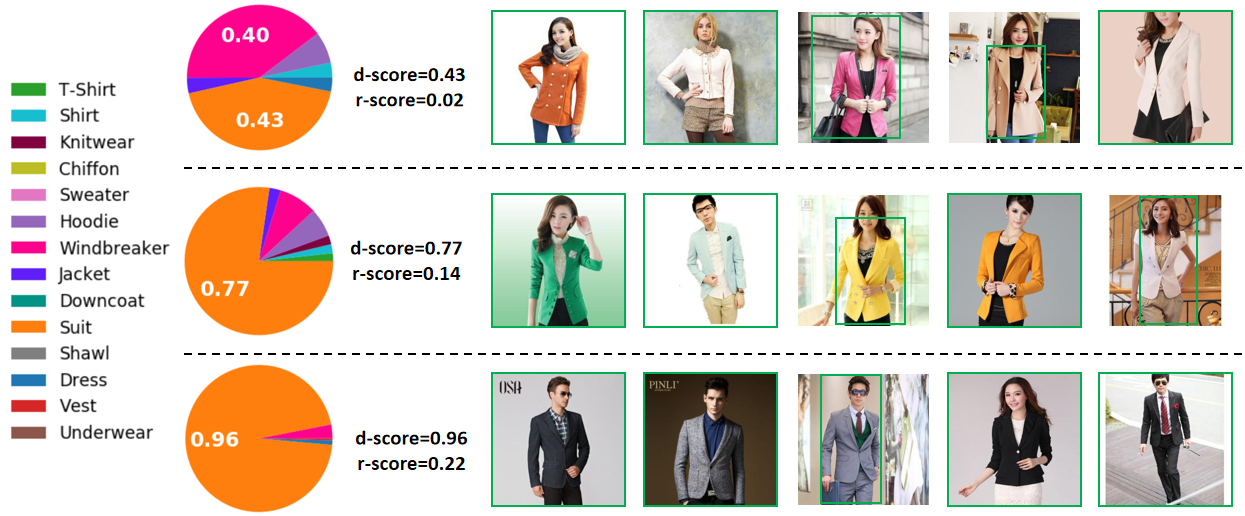}
    \caption{Visualization of our memory module on Clothing1m dataset. We exhibit three key slots with ``suit" as the category of interest, with each row standing for one key slot. The left pie chart shows the category distribution in each key slot, and the right images are representative ROIs belonging to each key slot. The d-score and r-score for ``suit" category of each key slot are also reported.}
    \label{fig:prototype}   
\end{figure*}
\subsection{Training Algorithm} For better representation, we use $\bm{\theta}_\text{cnn}$ to denote the model parameters of CNN and $\bm{\theta}_\text{mem}$ to denote $\{\mathbf{K},\mathbf{D},\mathbf{R}\}$ in memory module.  

At first, we utilize initial ROI weights $\bar{\texttt{w}}(\mathbf{x}_i)$ mentioned in Section~\ref{sec:weight_init} to obtain weighted average of ROI-level features as bag-level features, which are used to train  the CNN model $\bm{\theta}_\text{cnn}$  and the memory module $\bm{\theta}_\text{mem}$. Then, we train the whole system in an end-to-end manner. Specifically, we train the CNN model $\bm{\theta}_\text{cnn}$  and the memory module $\bm{\theta}_\text{mem}$ with the bag-level features $\bar{\mathbf{x} }= \sum_{\mathbf{x}_i\in\mathcal{B}} \texttt{w}(\mathbf{x}_i)\cdot \mathbf{x}_i$, while the weights of ROIs  $\texttt{w}(\mathbf{x}_i)$ are updated accordingly based on updated $\bm{\theta}_\text{cnn}$ and $\bm{\theta}_\text{mem}$.  
In this way, cleaner bag-level features can help learn better key slots and value slots in the memory module, while the enhanced memory module can assign more reliable ROI weights and contribute to cleaner bag-level features in turn.
The total loss of the whole system can be written as:
\begin{eqnarray} \label{L_total}
\mathcal{L}_\text{all}= \mathcal{L}_\text{cls}+\mathcal{L}_\text{memory}.
\end{eqnarray} 

For better performance, we leverage the idea of curriculum learning \cite{Curriculum-Learning}. It suggests that when training a model, we should start with clean or simple training samples to have a good initialization, and then add noisy or difficulty training samples gradually to improve the generalization ability of the model.
After calculating the ROI weights, the top-score ROIs in each bag should be cleaner than the low-score ones. So $p$ is used as a threshold parameter to filter out the noisy ROIs in each bag.
Following the idea of curriculum learning, $p$ is set to be relatively small at first so that the selected ROIs are very discriminative and representative. Then we increase $p$ gradually to enhance the generalization ability of the trained model.   
The total training process can be seen in Algorithm \ref{alg1}.

For evaluation, we directly use the well-trained CNN model to classify test images based on image-level features without extracting region proposals. The memory module is only used to denoise web data in the training stage, but not used in the testing stage.
 
\section{Experiments}
In this section, we introduce the experimental settings and demonstrate the performance of our proposed method.

\subsection{Datasets}
\noindent\textbf{Clothing1M:} 
Clothing1M \cite{Clothing1m} is a large-scale fashion dataset designed for webly supervised learning. It contains about one million clothing images crawled from the Internet, and the images are categorized into 14 categories. Most images are associated with noisy labels extracted from their surrounding texts and used as the training set. A few images with human-annotated clean labels are used as the clean dataset for evaluation.

\noindent\textbf{Food-101 \& Food-101N:}
Food-101 dataset \cite{food101} is a large food image dataset collected from \textit{foodspotting.com}.  It has 101 categories and 1k images for each category with human-annotated labels. 
Food-101N is a web dataset provided by \cite{CleanNet}.
It has 310k images crawled with the same taxonomy in Food101 from several websites (excluding \textit{foodspotting.com}).
In our experiments, we use Food-101N for training and Food-101 for evaluation.

\noindent\textbf{Webvision \& ILSVRC:}
The WebVision dataset \cite{WebVision} is composed of training, validation, and test set. The training set is crawled from Flickr and Google by using the same 1000 semantic concepts as in the  ILSVRC-2012 \cite{ILSVRC} dataset. It has 2.4 million images with noisy labels. The validation and test set are manually annotated. In our experiments, we only use the WebVision training set for training but perform the evaluation on both WebVision validation set (50k) and ILSVRC-2012 validation set (50k). 

\subsection{Implementation Details}\label{Implementation-Details}
We adopt ResNet50 \cite{resent} as the CNN model and use the output of its last convolutional layer as the feature map to extract ROI features. 
For Clothing1M and Food101N, we use ResNet50 pretrained on ImageNet following previous works \cite{CleanNet,CurriculumNet}. For WebVision and ImageNet, ResNet50 is trained from scratch with the web training images in WebVision.

For the proposal extractor (\emph{i.e.}, Edge Boxes), there are two important parameters $\texttt{MaxBoxes}$ and $\texttt{MinBoxArea}$, in which $\texttt{MaxBoxes}$ controls the maximal number of returned region proposals and $\texttt{MinBoxArea}$ determines the minimal area. In our experiments, we use $\texttt{MaxBoxes}=20$ (\emph{i.e.}, $n_p=20$) and $\texttt{MinBoxArea}=5000$. By default, we use two images in each training bag (\emph{i.e.}, $n_g=2$), so the number of ROIs in each bag is $n_b=n_g(n_p+1)=2\times(20+1)=42$.

\begin{figure}[t]
    \centering
    \includegraphics[width=8cm]{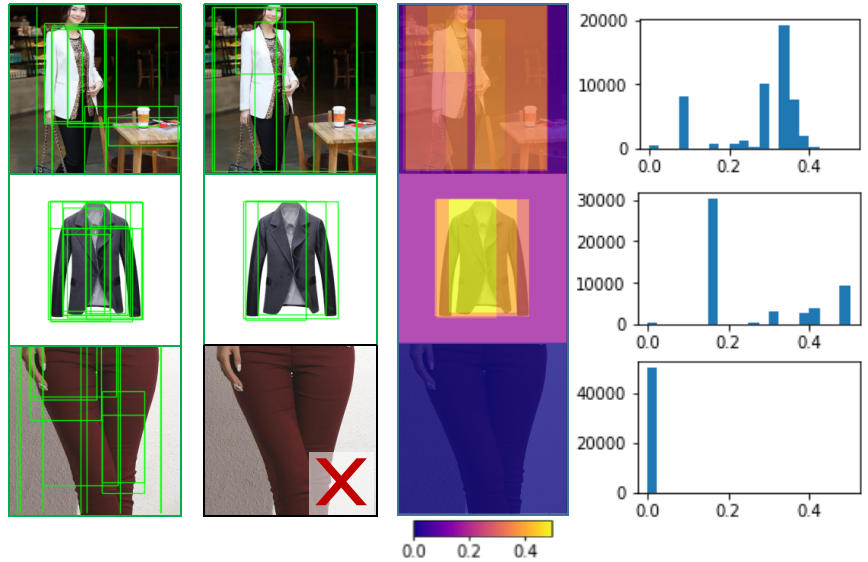}
    \caption{\textbf{Column-1:} A training bag for ``Suit" with $n_g=3$.  \textbf{Column-2:} Top 30\% ROIs selected by our network. The bottom image together with all its region proposals are removed. \textbf{Column-3\&4}: The heat map obtained by summing over ROI weights and the histogram of the heat map pixels.}
    \label{heatmap}   
\end{figure}

\subsection{Qualitative Analyses}
In this section, we provide in-depth qualitative analyses to elaborate on how our method works. We first explore the memory module and then explore training bags.

\noindent{\textbf{Memory Module: }}By taking Clothing1M dataset as an example and ``suit" as category of interest, we choose three key slots for illustration in Figure~\ref{fig:prototype}, in which each row stands for one key slot with its corresponding d-score and r-score. To visualize each key slot, we select five ROIs with highest cosine similarity to this key slot, \emph{i.e.}, $\text{cos}(\mathbf{x}_i,\mathbf{k}_l)$.

The first key slot almost clusters an equal number of bags from both ``Suit'' and ``Windbreaker'' as the pie chart shows, so it has the lowest d-score. In the meantime, its total number of bags from ``Suit'' is smaller than those of the other two slots, so its r-score is also the lowest. Hence, such a key slot is neither discriminative nor representative to ``Suit''. 

The other two key slots are very discriminative for the category ``Suit'' and have a very high d-score. However, the third key slot is more representative for ``Suit'' than the second one, resulting in a higher r-score. The explanation is that the total number of bags with colorful suits (second key slot) is smaller than those with black/grey suits (third key slot), so we claim that the third key slot is more representative to ``Suit". Combining the r-score and d-score, we would claim that the third key slot is the most prototypical to ``Suit" (see the visualization of all $L=144$ key slots in Supplementary).

\noindent{\textbf{Training Bags: }}
Based on the memory module, different ROIs within each bag are assigned with different weights, according to their areas and prototypical scores of their nearest key slots. 
In Figure \ref{heatmap}, we show a training bag with three images (\emph{i.e.}, $n_g=3$). By comparing the first and the second column, we can observe that the noisy images and noisy region proposals have been removed based on the learned ROI weights. By summing over the ROI weights, we can obtain the attention heat map with brighter colors indicating higher weights. The heat maps and their corresponding histograms are shown in the third and fourth columns, respectively.
It can be seen that the background region has lower weights than the main objects. Therefore, the results in Figure \ref{heatmap} demonstrate the ability of our network to address the label noise and background noise.

\subsection{Ablation Study}
\label{subsec:Ablation-Study}
We first compare the results of our method with different $n_g$ and $L$ in Table \ref{ablation-2}, by taking the Clothing1M dataset as an example. 

\noindent{\textbf{The bag size: }}As Table \ref{ablation-2} shows, our method with $n_g=2$ achieves the best performance, so we use it as the default parameter in the rest experiments. Furthermore, it can be seen that the performance with $n_g=1$ is worse than those with $n_g>1$, because our method with only one image in a bag will be unable to reduce the label noise when it is a noisy image. 

\noindent{\textbf{The number of key slots: }}As in Table \ref{ablation-2}, the performance of our method is quite robust when the number of key slots is big enough ($L \geq 8\times 8$), while a too-small number ($L = 4\times4$) will lead to a performance drop.
Notice that the best performance is achieved with $L = 12\times 12$ while the Clothing1M dataset has 14 categories, it suggests that when each category can occupy about $12\times 12\div 14 \approx 10$ clustering centers in the memory module, our method will generally achieve satisfactory performance. Following this observation, we have $L=12\times12$ for Clothing1M, $L=32\times32$ for Food101, and $L=100\times100$ for WebVision and ImageNet in the rest experiments.

\begin{table}[tb]
\begin{center}
\begin{tabular}{|l|c|c|c|c|c|c|c|}
\hline
\textbf{Parameter} $n_g$ & 1 & \textbf{2} & 3 & 4 & 5   \\ \hline
\textbf{Accuracy} & 72.9 & \textbf{82.1} & 81.1 & 78.5 & 75.9   \\ \hline
\textbf{Parameter} $L$ & $4^2$  & $8^2$  & $\mathbf{12^2}$  & $16^2$  & $20^2$  \\ \hline
\textbf{Accuracy} & 78.7  & 81.7 & \textbf{82.1} & 81.9 & 81.6 \\ \hline
\end{tabular}
\end{center}
\caption{Accuracies (\%) of our method with different $n_g$ and $L$ on the Clothing1M. The best result is denoted in boldface.}
\label{ablation-2}
\end{table}

Secondly, we study the contribution of each component in our method in Table~\ref{ablation}. 
ResNet50 and ResNet50+ROI are two naive baselines, and SOMNet denotes our method. ResNet50+ROI uses both images and region proposals as input, but it only achieves slight improvement over ResNet50, which shows that simply using proposals as data augmentation is not very effective.  

\setlength{\tabcolsep}{6pt}
\begin{table*}[tb]
	\begin{center}
		\begin{tabular}{|c|c|c|c|c|c|c|}
			\hline
			 ResNet50 & ResNet50+ROI & SOMNet (w/o d-score) & SOMNet (w/o r-score) & SOMNet (w/o a-score)   \\ \hline
			68.6 &  69.1 & 76.8 & 73.5 &74.8  \\ \hline
			SOMNet ($p=40\%$)  & SOMNet (w/o ROI)  &  SOMNet+K-means & SOMNet (w/o SOM) & \textbf{SOMNet} \\ \hline
			81.3   & 74.1 & 79.5& 78.2 &\textbf{82.1} \\ \hline
		\end{tabular}
	\end{center}
	\caption{Accuracies (\%) of our method and special cases on the Clothing1M. The best result is denoted in boldface.}
	\label{ablation}
\end{table*}

\setlength{\tabcolsep}{8pt}
\begin{table*}[tb]
\begin{center}
\begin{tabular}{|l|c|c|c|c|c|c|}
\hline
\multicolumn{1}{|c|}{\textbf{Training set}}  & \textbf{Clothing1M} & \textbf{Food101N} & \multicolumn{4}{c|}{\textbf{WebVision}} \\ 
\hline
\multicolumn{1}{|c|}{\textbf{Test set}}  & \textbf{Clothing1M} & \textbf{Food101} & \multicolumn{2}{c|}{\textbf{WebVision}} & \multicolumn{2}{c|}{\textbf{ImageNet}} \\ \hline
\multicolumn{1}{|c|}{\textbf{Evaluation metric}}                        & Top-1      & Top-1   & Top-1          & Top-5         & Top-1         & Top-5         \\ \hline
ResNet50                                   & 68.6       & 77.4    & 66.4           & 83.4          & 57.7          & 78.4          \\\hline
2014 Sukhbaatar \etal \cite{label-flip}  &    71.9 &      80.8   &          67.1     &  84.2             &  58.4             &    79.5           \\
2015 DRAE \cite{DRAE}      &     73.0       &     81.1    &         67.5&  85.1 &    58.6 &   79.4            \\ 
2017 Zhuang \etal \cite{Attend}                & 74.3       &     82.5    & 68.7           & 85.4          & 60.4  & 80.3      \\
2018 AD-MIL \cite{ilse2018attention}   & 71.1      & 79.2        &         66.9       &            84.0   &        58.0       &       78.9        \\
2018 Tanaka \etal \cite{bs-tanaka2018joint} &72.2*&	81.5&	67.4&	84.7&	59.5&	80.0 \\
2018 CurriculumNet \cite{CurriculumNet} & 79.8     & 84.5  & 70.7&	88.6&	62.7&	83.4\\
2019 SL \cite{ICCV-wang2019symmetric} & 71.0*&	80.9&	66.2&	82.3&	58.7&	78.8 \\
2019 MLNT \cite{bs-Li_2019_CVPR}& 73.5*&	82.5&	68.3&	85.0&	60.2&	80.1\\
2019 PENCIL \cite{bs-yi2019probabilistic}& 73.5* &	83.1&	68.9&	85.7&	60.8&	81.1\\
\hline
\textbf{SOMNet}                                      & \textbf{82.1}       & \textbf{87.5}    & \textbf{72.2}           & \textbf{89.5}         & \textbf{65.0}         &\textbf{85.1}         \\ 
\hline
\end{tabular}
\end{center}
\caption{The accuracies (\%) of different methods on Clothing1M, Food101, Webvision, and ImageNet datasets. The results directly copied from corresponding papers are marked with ``*''. The best results are denoted in boldface.}
\label{allresult}
\end{table*}

\noindent{\textbf{Three types of scores: }}Recall that we use three types of scores to weight ROIs: r-score, d-score, and a-score. To investigate their benefit, we report the performance of our method by ablating each type of score, denoted by SOMNet(w/o d-score), SOMNet(w/o r-score), and SOMNet(w/o a-score) in Table~\ref{ablation}. We observe that the performance will decrease with the absence of each type of score, which indicates the effectiveness of our designed scores. 

\noindent{\textbf{Curriculum learning: }}Notice we utilize the idea of curriculum learning by gradually increasing $p$ from 10\% to 40\% during training. In Table \ref{ablation}, SOMNet ($p=40\%$) denotes the result of directly using $p = 40\%$ from the start of training, which has a performance drop of 0.8\%. It proves the effectiveness of using curriculum learning.

\noindent{\textbf{Background noise removal: }}We claimed that web data have both label noise and background noise. To study the influence of background noise, we only handle label noise by not using region proposals, and the result is denoted by SOMNet (w/o ROI) in Table \ref{ablation}. To make a fair comparison, we find out that $n_g=5$ and $p=60\%$ are the optimal parameters in this setting. 
However, the best result is only $74.1\%$, which is much worse than using region proposals. This result demonstrates the necessity of handling background noise.

\noindent{\textbf{The self-organizing memory module: }}
Since traditional clustering methods like K-means can realize a similar function to the memory module, we replace our self-organizing memory module with the K-means method and refer to this baseline as SOMNet+K-means (see implementation details in Supplementary). The performance drop in this setting proves the effectiveness of joint optimization in an end-to-end manner with our memory module.
Moreover, to demonstrate the effectiveness of using SOM as an extension, we set neighborhood size as $1$, which is equivalent to removing SOM. The comparison between SOMNet (w/o SOM) and SOMNet indicates that it is beneficial to use SOM in our memory module. 

\subsection{Comparison with the State-of-the-Art}
We compare our method with the state-of-the-art webly or weakly supervised learning methods in Table \ref{allresult} on four benchmark datasets: Clothing1M, Food101, WebVision, and ImageNet.
The baseline methods include  Sukhbaatar \etal  \cite{label-flip}, DRAE \cite{DRAE}, Zhuang \etal  \cite{Attend}, AD-MIL \cite{ilse2018attention}, Tanaka \etal \cite{bs-tanaka2018joint}, CurriculumNet~\cite{CurriculumNet}, SL \cite{ICCV-wang2019symmetric}, MLNT \cite{bs-Li_2019_CVPR}, and PENCIL \cite{bs-yi2019probabilistic}. Some methods did not report their results on the above four datasets. Even with evaluation on the above datasets, different methods conduct experiments in different settings (\emph{e.g.}, backbone network, training set), so we re-run their released code in the same setting as our method for fair comparison. For those methods which already report results in exactly the same setting as ours, we directly copy their reported results (marked with ``*'').

From Table \ref{allresult}, we can observe that our method achieves significant improvement over the backbone ResNet50. The average relative improvement (Top-1) on all four datasets is 9.18\%. It also outperforms all the baselines, demonstrating the effectiveness of our method for handling label noise and background noise using the memory module.

\section{Conclusion}
In this paper, we have proposed a novel method, which can address the label noise and background noise of web data at the same time. Specifically, we have designed a novel memory module to remove noisy images and noisy region proposals under the multi-instance learning framework. Comprehensive experiments on four benchmark datasets have verified the effectiveness of our method. 

\section*{Acknowledgement}
The work is supported by the National Key R\&D Program of China (2018AAA0100704) and is partially sponsored by National Natural Science Foundation of China (Grant No.61902247) and Shanghai Sailing Program (19YF1424400).

{\small
\bibliographystyle{ieee_fullname}
\bibliography{egbib}
}

\end{document}